\documentclass[11pt]{article}

\usepackage[final]{acl}
\usepackage{enumitem}

\ifdefined\OspreyArxivStyle
\else
\usepackage{times}
\fi
\usepackage{latexsym}
\usepackage[T1]{fontenc}
\usepackage[utf8]{inputenc}
\usepackage{microtype}
\ifdefined\OspreyArxivStyle
\else
\usepackage{inconsolata}
\fi
\usepackage{amsmath}
\usepackage{amssymb}
\usepackage{booktabs}
\usepackage{multirow}
\usepackage{graphicx}
\usepackage{xcolor}

\graphicspath{{figures/}}

\newcommand{\target}{p_{\theta}}
\newcommand{\draft}{q_{\phi}}
\newcommand{\hidden}{\mathbf{h}}

\newcommand{\ctx}{\mathbf{c}}

\title{Osprey: Target-agnostic Pre-training Makes Stronger Drafters in Speculative Decoding}

\author{%
\textbf{Fengxiang Bie}\thanks{\raggedright Equal Contribution.
Correspondence to: Tianyi Zhang \texttt{<tonyzhang@together.ai>}.}%
\textsuperscript{1,2}, \textbf{Yuqing Jian\textsuperscript{*,1},
Yifan Yu\textsuperscript{1,4}, Zhongzhu Zhou\textsuperscript{1,2}, Zelei Shao\textsuperscript{1,4},} \\
\textbf{Ben Athiwaratkun\textsuperscript{1}, Shuaiwen Leon Song\textsuperscript{1,2},
Chenfeng Xu\textsuperscript{1,3}, Xiaoxia Wu\textsuperscript{1}, Tianyi Zhang\textsuperscript{1}} \\[0.5em]
\textsuperscript{1}Together AI \quad
\textsuperscript{2}The University of Sydney, Australia \\
\textsuperscript{3}The University of Texas at Austin, USA \quad
\textsuperscript{4}University of Illinois Urbana-Champaign, USA \\[0.4em]
\textit{Accepted at EMNLP 2026.}
}

\begin{document}
\maketitle

\begin{abstract}
Speculative decoding is critical for accelerating LLM inference. However, the speedup is fragile: drafters are typically trained against a narrow distribution for a single target model, and their acceptance rate collapses under workload shifts. This is a striking inversion of modern LLM development, where target models are valued precisely for the broad generalization they acquire through large-scale pretraining.
We argue that the natural remedy---pretraining---has been hard to apply to drafters because existing recipes are target-specific: the drafter consumes the target's hidden states and is distilled on the target's logits, so pretraining must be repeated for each target.
We introduce \textbf{Osprey}, which instead bootstraps drafters from off-the-shelf pretrained small language models, treating broad pretraining as a reusable, target-agnostic asset and reducing per-target work to a lightweight adaptation step. Realizing this requires overcoming two challenges: small LMs are far deeper than a latency-bound drafter can afford, and their pretrained computation must remain intact while the drafter learns to ingest target hidden states and emit tokens in the target's vocabulary. Osprey addresses both by pruning to a shallow backbone, restoring its language-modeling capability with target-agnostic next-token pretraining, and adapting it to each target through vocabulary alignment, zero-initialized QKV expansion, and distillation from the target model's output distribution.
Empirically, a single pretrained Osprey backbone transfers across targets and improves mean acceptance length by $16.1\%$ for Qwen3-8B, \textcolor{black}{$21.2\%$ for Llama-3.3-70B-Instruct,} and $22.7\%$ for the 229B MiniMax-M2.5 (with $17.5\%$ higher tokens per second), with the largest gains on out-of-domain and multilingual data. Our code is available at \url{https://github.com/LeanModels/Osprey}.

\end{abstract}

\section{Introduction}

Speculative decoding~\citep{leviathan2023speculative,chen2023accelerating,li2025eagle3,cai2024medusa} has emerged as one of the most effective techniques for accelerating LLM serving. A lightweight \emph{drafter} proposes a block of future tokens, while a larger \emph{target} model verifies these tokens in parallel. Since accepted tokens are committed exactly according to the target distribution and rejected positions fall back to standard autoregressive decoding, speculative decoding provides a rare combination of acceleration and losslessness. This has made it particularly attractive for latency-sensitive and resource constrained deployment settings, where prior systems commonly report $2$--$5\times$ decoding speedups~\citep{sd-survey}.

Yet this success hides a fundamental fragility. In practice, the drafter is often trained or tuned on a particular data distribution, model family, or deployment domain~\citep{sd-domain-shift}. When the serving workload shifts, for example, from chat to code, from general QA to domain-specific reasoning, or from short responses to long-form generation, the drafter's predictions can become poorly aligned with the target model. The result is a sharp drop in acceptance rate, and therefore a sharp loss of the very speedup that speculative decoding is designed to provide. As we show in Figure~\ref{fig:main-comparison}, a state-of-the-art Eagle3~\citep{li2025eagle3} drafter that is trained on a certain distribution of data performs well on in-domain data while performing noticeably worse on out-of-domain data (e.g. Eagle3 trained on math achieves an average of 5.06 acceptance length on math data, and an average of only 1.86 acceptance length on code data).

This brittleness is interesting in the context of today's LLM development~\citep{scaling-laws}. The target models we serve today are trained on increasingly broad mixtures of data and are valued precisely for their ability to generalize across domains, tasks, and interaction styles~\cite{understanding-gen-pretraining}. Domain shift, once a central concern in classical machine learning, is rarely a dominant narrative for frontier LLMs; the LLM field instead emphasizes emergent capabilities and broad transfer~\cite{emergent-abilities}. However, the drafter used to accelerate these models remains a much more narrow and fragile component. In other words, speculative decoding inherits the latency benefits of small models, but not the generalization benefits of the large models it accelerates.

This motivates us to rethink the design of the drafter. If the fragility of speculative decoding comes from distribution shift, then a natural question is whether we can improve the drafter's generalization in the same way machine learning has historically improved generalization, such as through pretraining. Indeed, the role of pretraining in learning transferable representations has been studied for decades~\citep{erhan2010does, understanding-gen-pretraining, bert}. One might therefore try to pretrain existing speculative drafters, such as EAGLE-style drafters on more diverse data. However, these methods are inherently target-specific: the drafter consumes activations from the target model and is trained in the target's representation space. Consequently, adapting the method to a new target model requires a new, expensive training pipeline.  More importantly, such a training pipeline still does not directly inherit the broad generalization ability already learned by existing pretrained language models.

So we suggest a different direction. Today, small language models (SLMs)~\cite{slm} are increasingly capable, broadly pretrained, and close in scale to the draft models used in speculative decoding. Unlike target-specific drafters trained from scratch, these models already encode general linguistic, reasoning, and domain knowledge from large-scale pretraining. The question is therefore: can speculative decoding use pretrained small LMs as general-purpose drafters, while still aligning them tightly enough with the target model to preserve high acceptance rates?



Building the speculators on top of small language models, however, is non-trivial. Existing small language models (SLMs) are typically 16 or more layers deep~\cite{tinyllama}. A drafter must be shallow, as a deep drafter would negate the latency benefits of speculation. Furthermore, the backbone needs to be compatible with a wide range of target models. Its embedding and LM head must be adaptable to any target tokenizer, and it must be able to ingest target hidden states without disrupting pretrained capabilities.

\begin{figure*}[t!]
    \centering
    \includegraphics[width=\textwidth]{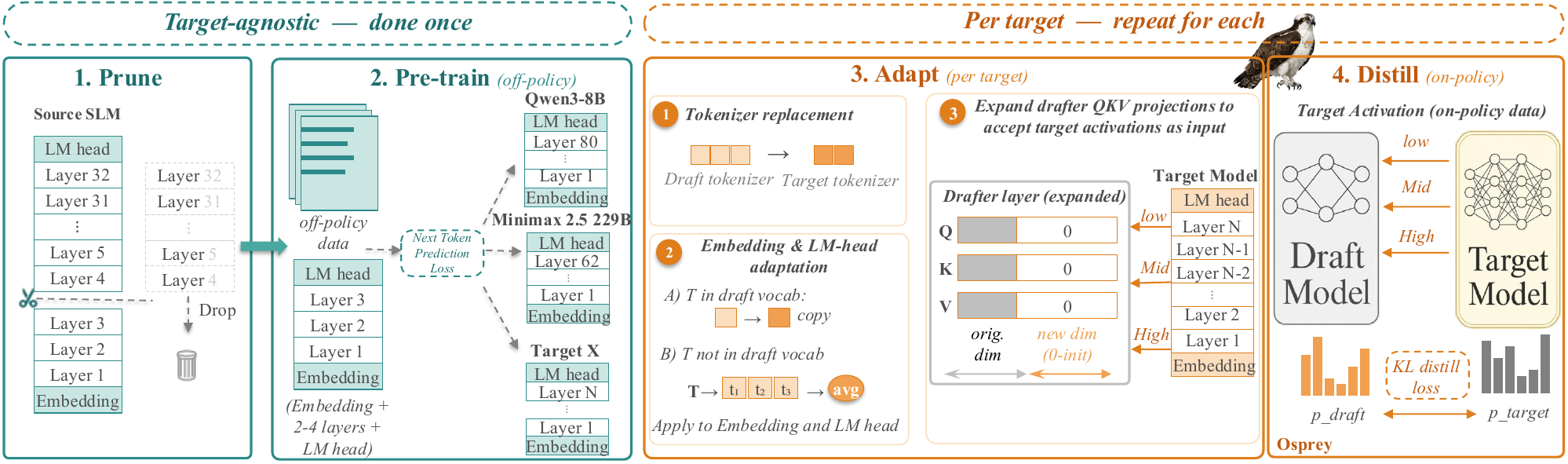}
    \caption{Overview of \textbf{Osprey}. The pipeline has four stages, split into a target-agnostic phase done once (left) and a per-target phase repeated for each new target model (right). \textbf{(1) Prune}: keep only the embedding, LM head, and the first few transformer layers of an off-the-shelf small LM. \textbf{(2) Pre-train (off-policy)}: train the pruned model with next-token prediction on a generic corpus to restore language-modeling capability. \textbf{(3) Adapt (per target)}: replace the tokenizer and align the embedding/LM head with the target's vocabulary, then expand the QKV projections with zero-initialized taps so the drafter can ingest target hidden states without disturbing the pretrained initialization. \textbf{(4) Distill (on-policy)}: distill from target activations and output distributions via KL on target-generated sequences. The same target-agnostic checkpoint is reused across targets ranging from Qwen3-8B to MiniMax-M2.5-229B.}
    \label{fig:overview}
\end{figure*}

To address these challenges, we propose a four-stage training recipe for speculative drafters (Figure~\ref{fig:overview}), named \textbf{Osprey}, that is split into a universal, target-agnostic phase and a per-target adaptation phase. First, we \emph{prune} an off-the-shelf pretrained small language model down to its embedding, language modeling head, and first few transformer layers to construct a shallow backbone. Second, we \emph{pre-train} this pruned model off-policy via next-token prediction on a large, generic corpus, which restores robust and target-independent language modeling capabilities. \textcolor{black}{Crucially, this pretraining stage is entirely \emph{target-agnostic}, yielding a drafter backbone that can be reused across different target models.} Third, we \emph{adapt} this universal checkpoint to a specific target model by replacing the tokenizer and aligning the embedding and language modeling head with the target's vocabulary. During this adaptation step, we also expand the QKV projections with zero-initialized columns, allowing the drafter to ingest target hidden states without disrupting the pretrained initialization. Finally, we \emph{distill} target-specific behavior into the drafter by conditioning it on target activations and training it to predict target logits. \textcolor{black}{Thus, each deployed drafter is ultimately target-specific, and inherits the broad language prior of the reusable backbone.} This decoupled approach allows a single target-agnostic checkpoint to be efficiently reused across a wide range of architectures, from Qwen3~\citep{qwen3}, through Llama-3.3~\citep{llama3.3}, to MiniMax-M2.5~\cite{minimaxm25}.

In this work, our primary contributions are twofold. We first propose a \emph{target-agnostic pretraining} recipe for speculative drafters that disentangles general language modeling from target conditioning, creating a single, highly reusable drafter backbone. Second, we demonstrate that Osprey delivers higher acceptance rates and superior speedups than \textcolor{black}{strong EAGLE-3 baselines}. Empirically, on Qwen3-8B, Osprey improves the mean acceptance length over EAGLE-3 across all five evaluation domains (chat, code, commonsense, finance, and math), with the most significant gains at further lookahead positions. \textcolor{black}{For larger targets, Osprey improves mean acceptance length and tokens per second over matched EAGLE-3 \citep{li2025eagle3} baselines by $21.2\%$ and $17.9\%$ for Llama-3.3-70B-Instruct, and by $22.7\%$ and $17.5\%$ for MiniMax-M2.5.} Furthermore, Osprey exhibits excellent transferability to multilingual data (such as 42-language Global-MMLU and 11-language MGSM benchmarks) despite using only code data for target-specific training.

\section{Related Work}
\label{sec:related}


\paragraph{Speculative decoding and target-conditioned drafters.}
Speculative decoding accelerates autoregressive inference by using a cheaper drafter to propose multiple tokens that the target model verifies in parallel \citep{leviathan2023speculative,chen2023accelerating}. Existing drafters range from external proposal models \citep{xia2023specdec,miao2024specinfer} to target-conditioned architectures that consume target hidden states, including EAGLE \citep{li2024eagle}, EAGLE-2 \citep{li2024eagle2}, and EAGLE-3 \citep{li2025eagle3}, as well as Medusa, Hydra, ReDrafter, Clover, P-EAGLE, Falcon, and HASS \citep{cai2024medusa,ankner2024hydra,bhendawade2024redrafter,xiao2024clover,xiao2024clover2,hui2026peagle,gao2024falcon,zhang2024hass}. A parallel line of work avoids a separate drafter by reusing the target itself, through early exits, adaptive layer skipping, adapters, or multi-stream attention \citep{elhoushi2024layerskip,zhang2024draftverify,xia2024swift,liu2024kangaroo,bhendawade2024specstreaming}. These methods reduce inference cost, but either train drafters from scratch, tightly couple drafting behavior to a specific target, or require target-side architectural/training changes. Osprey instead keeps the target untouched and builds a standalone shallow drafter whose body is initialized from a target-agnostic pre-trained next-token predictor. We adopt EAGLE-3's TTT paradigm and the three-layer target hidden states tap, but inject target features through a zero-initialized QKV expansion, preserving the pre-trained drafter function at adaptation.

\paragraph{Robust and reusable draft-model training.}
Recent studies show that speculative acceptance can vary sharply across domains and user populations \citep{sd-survey,sandler2025disparate,sd-domain-shift}, motivating drafters that generalize beyond the target-specific distillation distribution. Prior work improves robustness through pre-trained SLM initialization, state-space drafters, online updates, cross-vocabulary verification, or universal verification rules \citep{berdoz2025steering,kim2025mamba,chen2025omnidraft,timor2025universal,liu2024osd}. Osprey differs by explicitly pre-training a reusable shallow LM on generic web text \citep{penedo2024fineweb} and then lightly adapting it to each target, rather than relying on online correction or selecting among existing drafters. Our contribution is not a new distillation objective but a reusable initialization and adaptation pathway that yields stronger domain robustness and cross-target transfer under the same speculative decoding interface. A more complete discussion of related sub-fields---depth pruning, vocabulary transfer, continued pre-training, white-box distillation, and drafter-free parallel decoding---is deferred to Appendix~\ref{sec:related-more}.

\section{Method}

Osprey constructs a target-conditioned speculative drafter through a four-stage pipeline organized into two phases (Figure~\ref{fig:overview}). In the target-agnostic phase, we prune a pretrained source language model to a shallow architecture (Stage 1) and continue next-token pretraining on generic text (Stage 2). These stages produce a reusable shallow language-modeling backbone without being coupled to a target model. In the target-specific phase, we align the backbone with the target vocabulary and augment it to use target hidden states (Stage 3). We then adapt the converted drafter to the frozen target model using distillation (Stage 4).

\subsection{Setup}

Let $p_\theta$ denote the frozen target model and $q_\phi$ the lightweight drafter. Given a prefix $x_{\leq t}$, the drafter proposes $K$ draft tokens, which the target verifies in a single forward pass using the standard speculative-decoding acceptance rule~\citep{leviathan2023speculative}. Osprey changes how the drafter is constructed and trained; it does not modify the verification procedure.

The source model used to initialize $q_\phi$ may belong to a different model family and may use a different tokenizer from $p_\theta$. Stages 3--4 therefore align the drafter with the target vocabulary, hidden states, and output distribution.

\subsection{Drafter Pretraining (Stages 1--2)}
\label{sec:method:pretraining}

The first two stages produce a shallow next-token-prediction (NTP) backbone that can be reused across target models.

\paragraph{Stage 1: Pruning.}
Starting from a source language model with $L$ transformer blocks, we retain only its first $N$ blocks together with its input embeddings, final normalization layer, and language-modeling head. We keep $N$ small ($N \in \{1,2,3,4\}$) because the drafter is executed autoregressively and its latency directly affects the speedups of speculative decoding.

For example, retaining two blocks from \texttt{Qwen3-4B} \citep{qwen3} produces a 2-layer, approximately 591M-parameter backbone, while retaining four blocks from \texttt{layerskip-llama3.2-1B} \citep{elhoushi2024layerskip} produces a 4-layer, approximately 506M-parameter backbone. Aggressive pruning can severely degrade the model's next-token prediction capability, so we continue pretraining the pruned model in the next stage to restore this capability. Early-exit-trained models, such as LayerSkip \citep{elhoushi2024layerskip}, provide a stronger initialization because auxiliary early-exit objectives explicitly train intermediate layers to predict the next token.


\paragraph{Stage 2: Pretraining.}

We continue pretraining the shallow backbone with the standard next-token prediction objective. In our experiments, we use a 100B-token sample from \texttt{FineWeb} \citep{penedo2024fineweb} and train for five epochs, corresponding to approximately 500B training tokens. This continued pretraining restores the backbone's next-token prediction capability after pruning and provides a broadly trained initialization for target-specific adaptation.


We intentionally use more training tokens than the compute-optimal allocation for a model of this size \citep{hoffmann2022chinchilla}. Our goal at this stage is to obtain a broadly trained language backbone for subsequent target-specific adaptation, rather than to minimize the cost of pretraining. We refer to the resulting model as the \emph{shallow NTP backbone}.

\subsection{Drafter Conversion (Stage 3)}
\label{sec:method:conversion}

Stage 3 converts the standalone shallow NTP backbone into a target-conditioned drafter. The conversion involves three components: projecting the target hidden states, expanding the attention projections with zero-initialized columns, and aligning the vocabulary. The drafter's MLP blocks, normalization layers, and other pretrained components are transferred without architectural changes.

\paragraph{Multi-layer target hidden states.}
Following EAGLE-3 \citep{li2025eagle3}, we condition the drafter on hidden states from low, middle, and high layers of the target model. At position $t$, we concatenate the three target hidden states and project them into the drafter's hidden space:
\begin{equation}
    \ctx_t
    =
    W_c
    \left[
        \hidden_t^{(\ell_{\mathrm{low}})};
        \hidden_t^{(\ell_{\mathrm{mid}})};
        \hidden_t^{(\ell_{\mathrm{high}})}
    \right].
\end{equation}
Here, $\ell_{\mathrm{low}}$, $\ell_{\mathrm{mid}}$, and $\ell_{\mathrm{high}}$ identify the selected target layers, and $W_c$ is a newly initialized linear projection. The resulting representation $\ctx_t$ is provided as input to every attention layer of the drafter.

\paragraph{Zero-initialized attention expansion.}
At the beginning of a drafting step, the current token $x_t$ has been selected, but the target model has not yet computed hidden states for that token. The latest available projected representation is therefore $\ctx_{t-1}$.

For a given drafter layer, let $\mathbf{z}_t$ denote the token-side representation that its pretrained query, key, and value projections would normally process at position $t$. We concatenate $\ctx_{t-1}$ with $\mathbf{z}_t$ and expand each pretrained attention QKV projection $W_\mathrm{qkv}$ with zero-initialized columns:
\begin{equation}
    \widetilde{W}_\mathrm{qkv}=[0\;\;W_\mathrm{qkv}].
\end{equation}
At initialization,
\begin{equation}
    \widetilde{W}_\mathrm{qkv}
    \begin{bmatrix}
        \ctx_{t-1}\\
        \mathbf{z}_t
    \end{bmatrix}
    =
    W_\mathrm{qkv}\mathbf{z}_t.
\end{equation}
Because the new columns are initialized to zero, the target hidden states initially do not affect the drafter's query, key, or value vectors. The attention layers therefore retain their target-agnostic pretrained behavior at initialization. During training, these columns and all other drafter weights are updated, allowing the drafter to use the target hidden states to improve draft-token predictions.


\paragraph{Vocabulary alignment.}
Let $\mathcal V_S$ and $\mathcal V_T$ denote the source and target vocabularies, respectively. We adapt the drafter to use the target tokenizer and vocabulary while retaining information from its own pretrained parameters. Specifically, we resize the drafter's input embedding and language modeling head to $\lvert\mathcal V_T\rvert$ and initialize the resized matrices from the drafter's pretrained weights rather than from the target model \citep{fast_vocab_transfer}. The resulting drafter can predict tokens in $\mathcal V_T$ while preserving knowledge acquired during pretraining.

For each token in the target vocabulary, the initialization depends on whether the same token is present in the source vocabulary. If it is, we directly copy the corresponding pretrained row. Otherwise, we convert the target token to text, tokenize that text using the source tokenizer, and initialize the new row as the mean of the corresponding source-token rows. We use the same procedure for both the input embedding and the language modeling head.

Formally, let $W_S$ denote either pretrained source matrix and let $\widetilde W$ denote its resized counterpart. For each target token $v \in \mathcal V_T \setminus \mathcal V_S$, let $(s_1,\ldots,s_{m_v})=\tau_S(\operatorname{str}(v))$ be the sequence obtained by tokenizing its string representation with the source tokenizer. We initialize each row as
\begin{equation}
    \widetilde W[v]
    =
    \begin{cases}
        W_S[v],
        & v \in \mathcal V_S, \\[3pt]
        \displaystyle
        \frac{1}{m_v}\sum_{i=1}^{m_v} W_S[s_i],
        & v \notin \mathcal V_S.
    \end{cases}
\end{equation}
Thus, shared tokens retain their pretrained rows, whereas target-only tokens are initialized from the mean of their source-token representations.

\subsection{Target Adaptation (Stage 4)}
\label{sec:method:adaptation}

After conversion, we train the drafter $q_\phi$ by distilling the frozen target $p_\theta$. All drafter parameters, both pretrained and newly added, are trained. We use the Training-Time Test (TTT) framework from EAGLE-3 \citep{li2025eagle3} to align training with autoregressive inference.

For each training sample, the target first produces logits as distillation targets and hidden states as drafter inputs. The drafter is then unrolled for $K_{\mathrm{train}}$ steps. At position $t$, it receives the teacher-forced token $x_t$ and the projected target representation $\ctx_{t-1}$. It produces its own hidden state and next-token distribution, and this hidden state becomes part of its autoregressive context at the next step. Thus, the input tokens are teacher-forced, while the drafter's hidden-state trajectory is generated autoregressively as it is during inference.

Let $p_\theta^{(k)}$ and $q_\phi^{(k)}$ denote the target's and drafter's next-token distributions at TTT step $k$. We train with the following discounted distillation loss, which emphasizes earlier drafting steps:
\begin{equation}
    \mathcal{L}_{\mathrm{adapt}}
    =
    \sum_{k=0}^{K_{\mathrm{train}}-1}
    \lambda^k
    \operatorname{CE}
    \left(
        p_\theta^{(k)},
        q_\phi^{(k)}
    \right),
    \quad
    \lambda=0.8.
\end{equation}

\section{Experiments}
\label{sec:experiments}

\begin{figure*}[t]
  \centering
  \includegraphics[width=\textwidth]{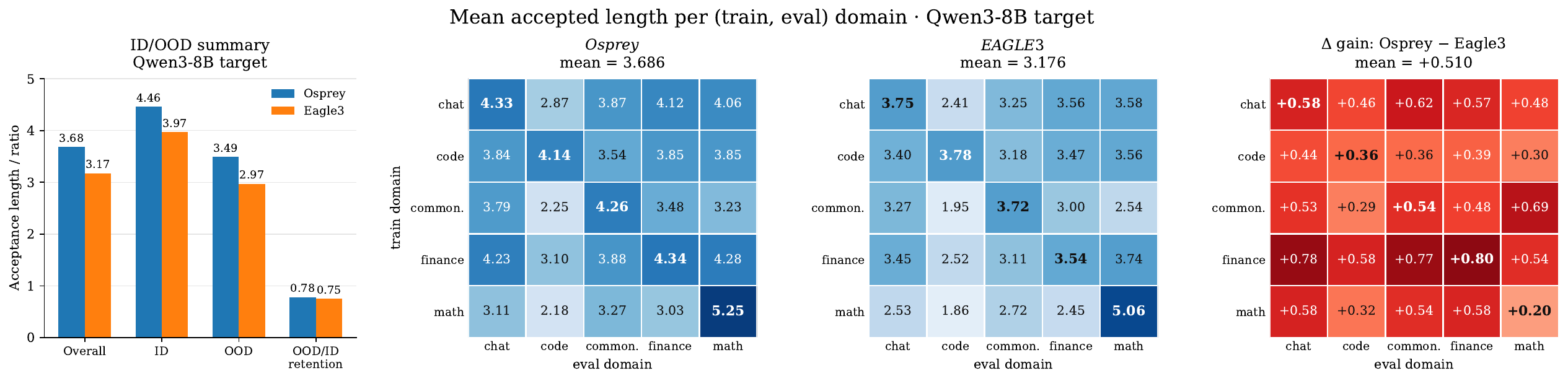}
  \caption{Mean accepted length (AL) for Qwen3-8B drafters adapted using data from one of five domains and evaluated across all five domains. Left: aggregate performance; middle: cross-domain AL matrices for Osprey and EAGLE-3 trained from scratch; right: cell-wise AL difference between Osprey and EAGLE-3 (Osprey minus EAGLE-3).}
  \label{fig:main-comparison}
\end{figure*}

We conduct comprehensive experiments comparing Osprey with competitive baselines. We first perform target-agnostic pretraining for a shallow next-token-prediction (NTP) backbone obtained by pruning Qwen3-4B to two layers. We then adapt Osprey drafters to three target models from different model families ranging from 8B to 230B parameters. Our experiments investigate whether target-agnostic pretraining improves end-to-end speedup across model scales, enhances out-of-domain (OOD) generalization, and provides sufficient gains to justify its computational cost.

\paragraph{Baselines, target models, and data.}
We compare Osprey with EAGLE-3~\citep{li2025eagle3}, a competitive speculative decoding drafter, on three target models: Qwen3-8B, Llama-3.3-70B-Instruct, and MiniMax-M2.5. For Qwen3-8B, we perform five separate adaptation runs for each method, with each run using data from one of five domains: chat, code, commonsense, finance, and math~\citep{onlinesd2024}. For each domain, we collect approximately $65$k problem prompts, generate responses with Qwen3-8B, and use the resulting prompt--response pairs to adapt the drafter. We then measure the mean accepted length (AL) of each adapted drafter on $512$ held-out prompts from each of the five domains. This $5 \times 5$ evaluation setup enables a controlled comparison of in-domain performance, where the adaptation and evaluation domains match, and OOD generalization, where they differ.

We further scale Osprey to two larger target models: the dense Llama-3.3-70B-Instruct and the mixture-of-experts (MoE) MiniMax-M2.5. For Llama-3.3-70B-Instruct, we compare Osprey against both the official checkpoint released by the EAGLE-3 authors and our reproduction of EAGLE-3. To ensure a controlled comparison, our reproduction and Osprey are adapted using the same target-regenerated Open-PerfectBlend dataset~\citep{perfectblend}. For MiniMax-M2.5, we adapt Osprey using $70$k coding prompts paired with responses generated by the target model~\citep{onlinesd2024}. We evaluate the adapted drafter on public benchmarks spanning mathematics, coding, chat, commonsense reasoning, and multilingual understanding: \textsc{MATH-500}~\citep{hendrycks2021math}, \textsc{HumanEval}~\citep{chen2021codex}, \textsc{LiveCodeBench}~\citep{jain2024livecodebench}, \textsc{MT-Bench}~\citep{zheng2023mtbench}, \textsc{Commonsense-Eval}, \textsc{MGSM}~\citep{shi2023mgsm}, and \textsc{Global-MMLU}~\citep{singh2025globalmmlu}.

\paragraph{Drafter training settings.}
All Osprey models are initialized from the same shallow NTP backbone, obtained by pruning Qwen3-4B to two layers. The shallow NTP backbone is then pretrained on FineWeb-100BT~\citep{penedo2024fineweb} for $55$k steps (approximately $462$B tokens), using a sequence length of $4096$ and a peak learning rate of $10^{-4}$. Complete pretraining details are provided in Appendix~\ref{sec:appendix_pretraining}.

For Qwen3-8B, Osprey and a one-layer EAGLE-3 baseline are adapted for $97.5$k steps using a TTT length of $5$, a sequence length of $4096$, and a learning rate of $10^{-4}$. For Llama-3.3-70B-Instruct, Osprey and the EAGLE-3 baseline (trained from scratch) use the same data and are each adapted for $37.5$k steps. We additionally evaluate the official EAGLE-3 checkpoint.\footnote{\url{https://huggingface.co/yuhuili/EAGLE3-LLaMA3.3-Instruct-70B}} Osprey is adapted to MiniMax-M2.5 for $35$k steps and compared with the publicly available one-layer EAGLE-3 checkpoint~\citep{thoughtworks2025eagle3}. Draft-model sizes and per-target training costs are reported in Appendix~\ref{app:draft-sizes}. We also apply target-agnostic pretraining to a four-layer LayerSkip Llama 8B model used as a DFlash drafter~\citep{chen2026dflash}; the results are presented in Appendix~\ref{app:dflash}.

\begin{table}[t]
\centering
\footnotesize
\setlength{\tabcolsep}{4pt}
\renewcommand{\arraystretch}{1.05}
\caption{Mean accepted length (AL) by evaluation domain and overall decoding throughput (tokens/s) for \texttt{Llama-3.3-70B-Instruct}.}
\label{tab:llama70b-results}
\begin{tabular}{@{}lccc@{}}
\toprule
Domain/metric & \shortstack{EAGLE-3\\reproduced} &
\shortstack{EAGLE-3\\official} & \textbf{Osprey} \\
\midrule
Chat        & 1.95  & 2.03  & \textbf{2.33} \\
Code        & 2.38  & 2.53  & \textbf{3.12} \\
CSense      & 2.31  & 2.33  & \textbf{2.55} \\
Finance     & 2.21  & 2.26  & \textbf{2.63} \\
Math        & 3.41  & 2.47  & \textbf{4.25} \\
Mean AL     & 2.45  & 2.33  & \textbf{2.97} \\
Throughput  & 155.8 & 150.3 & \textbf{183.7} \\
\bottomrule
\end{tabular}
\end{table}

\paragraph{Inference settings and metrics.}
We evaluate all drafters in SGLang~\citep{zheng2024sglang} with \texttt{batch=1}, \texttt{steps=5}, \texttt{topk=1}, and \texttt{draft\_tokens=6}. Results for larger batch sizes and different numbers of lookahead tokens, together with additional serving-sensitivity analyses, are provided in Appendix~\ref{app:compute}. MiniMax-M2.5 is served in FP8 with tensor-parallelism and expert-parallelism degrees of four.

We report mean accepted length (AL), defined as the average number of draft tokens accepted per verification step, and end-to-end decoding throughput in tokens per second (TPS). For Qwen3-8B, we additionally report mean in-domain (ID) and out-of-domain (OOD) performance, along with the OOD-to-ID ratio.

\subsection{Cross-Domain Evaluation on Qwen3-8B}
\label{sec:results:qwen3}

Figure~\ref{fig:main-comparison} compares Osprey with EAGLE-3 for Qwen3-8B, with each drafter adapted on one of five domains and evaluated on all five domains. Averaged over all $25$ combinations of adaptation and evaluation domains, Osprey achieves a mean AL of $3.686$, which is $16.1\%$ higher than EAGLE-3's $3.176$. Osprey achieves higher AL for every domain pair.

Osprey's advantage over EAGLE-3 is larger when the adaptation and evaluation domains differ. On OOD evaluations, Osprey achieves a mean AL of $3.493$, outperforming EAGLE-3's $2.977$ by $17.3\%$. On in-domain evaluations, Osprey achieves $4.465$, compared with $3.971$ for EAGLE-3, a $12.4\%$ advantage. The OOD-to-ID AL ratio is also higher for Osprey ($0.782$) than for EAGLE-3 ($0.750$). For the math-adapted drafters, Osprey achieves an in-domain AL of $5.25$ versus $5.06$ for EAGLE-3, and a mean AL of $2.90$ versus $2.39$ across the other four domains.

\subsection{Transfer to Larger Target Models}
\label{sec:results:larger-targets}

We next investigate whether the same pretrained shallow NTP backbone, obtained by pruning Qwen3-4B to two layers, can be adapted to target models outside the Qwen family. These target models test whether the pretrained backbone transfers despite differences in model family, architecture (dense versus mixture-of-experts), tokenizer, and hidden-state dimensionality.

\paragraph{Results on Llama-3.3-70B-Instruct.}
\label{sec:results:llama70b}

We adapt the pretrained two-layer Qwen3-4B checkpoint to \texttt{Llama-3.3-70B-Instruct}~\citep{llama3.3}. We compare Osprey with our reproduced EAGLE-3 baseline and the official EAGLE-3 checkpoint. As shown in Table~\ref{tab:llama70b-results}, Osprey achieves the highest AL in all five evaluation domains. Its mean AL and throughput are respectively $21.2\%$ and $17.9\%$ higher than those of our reproduced EAGLE-3 baseline, and $27.5\%$ and $22.2\%$ higher than those of the official EAGLE-3 checkpoint.

\begin{table}[t]
\centering
\caption{Accepted length (AL) and throughput (T/s) on five MiniMax-M2.5 public benchmarks, with $64$ prompts per benchmark.}
\label{tab:minimax-public-bench}
\small
\setlength{\tabcolsep}{4pt}
\renewcommand{\arraystretch}{1.15}
\begin{tabular}{@{}lcccc@{}}
\toprule
 & \multicolumn{2}{c}{EAGLE-3} & \multicolumn{2}{c}{Osprey} \\
\cmidrule(lr){2-3} \cmidrule(lr){4-5}
Benchmark & AL & T/s & AL & T/s \\
\midrule
\textsc{HumanEval}   & 2.570 & 184.4 & \textbf{3.565} & \textbf{243.2} \\
\textsc{MATH-500}    & 2.924 & 206.2 & \textbf{3.642} & \textbf{250.0} \\
\textsc{LiveCodeBench} & 2.508 & 176.2 & \textbf{3.779} & \textbf{249.1} \\
\textsc{MT-Bench}    & \textbf{2.868} & \textbf{214.4} & 2.805 & 201.6 \\
\textsc{Common.-Eval} & 2.302 & 154.3 & \textbf{2.367} & \textbf{154.8} \\
\midrule
Mean                 & 2.634 & 187.1 & \textbf{3.232} & \textbf{219.8} \\
\bottomrule
\end{tabular}
\end{table}

\paragraph{Results on MiniMax-M2.5.}
\label{sec:results:minimax}

We also adapt the same pretrained checkpoint to \texttt{MiniMax-M2.5}, a $229$B-parameter mixture-of-experts model served in FP8. 
Table~\ref{tab:minimax-public-bench} compares Osprey with the public EAGLE-3 checkpoint on the benchmarks described above. Osprey achieves a mean AL of $3.232$, which is $22.7\%$ higher than EAGLE-3's $2.634$, and a throughput of $219.8$ TPS, which is $17.5\%$ higher than EAGLE-3's $187.1$ TPS. Osprey's largest relative AL advantages occur on \textsc{LiveCodeBench} ($50.7\%$), \textsc{HumanEval} ($38.7\%$), and \textsc{MATH-500} ($24.6\%$). The two drafters perform similarly on \textsc{MT-Bench} and \textsc{Commonsense-Eval}.

\begin{table}[t]
\centering
\caption{Accepted length (AL) and throughput (T/s) on multilingual MiniMax-M2.5 benchmarks. MGSM uses $64$ prompts; each G-MMLU language setting uses $84$.}
\label{tab:minimax-multilingual}
\small
\setlength{\tabcolsep}{4pt}
\renewcommand{\arraystretch}{1.15}
\begin{tabular}{@{}lcccc@{}}
\toprule
 & \multicolumn{2}{c}{EAGLE-3} & \multicolumn{2}{c}{Osprey} \\
\cmidrule(lr){2-3} \cmidrule(lr){4-5}
Suite & AL & T/s & AL & T/s \\
\midrule
\textsc{MGSM} ($11$ langs)        & 2.275 & 167.3 & \textbf{2.557} & \textbf{179.7} \\
\textsc{G-MMLU} ($42$ langs)      & 2.633 & 197.5 & \textbf{3.259} & \textbf{231.0} \\
\textsc{G-MMLU} native            & 2.620 & 193.7 & \textbf{3.023} & \textbf{209.9} \\
\bottomrule
\end{tabular}
\end{table}

\paragraph{Multilingual benchmarks.}
\label{sec:results:minimax:multilingual}

Finally, we evaluate Osprey's OOD generalization beyond its coding adaptation data on the multilingual \textsc{MGSM} and \textsc{Global-MMLU} benchmarks and compare it with the public EAGLE-3 checkpoint. As shown in Table~\ref{tab:minimax-multilingual}, Osprey's AL is $12.4\%$ higher than EAGLE-3's on \textsc{MGSM} and $23.8\%$ higher on \textsc{Global-MMLU}. When the model must answer in the language of the question, Osprey's AL is $15.4\%$ higher than EAGLE-3's.

\subsection{Ablations}
\label{sec:ablations}

We conduct ablations using \texttt{Qwen3-8B} as the target model. Specifically, we evaluate the learning-rate robustness of Osprey and EAGLE-3, isolate the effect of target-agnostic pretraining, compare against a parameter-matched from-scratch baseline, and analyze when serving gains amortize the upfront pretraining cost. Unless otherwise stated, we vary one factor at a time while holding all other hyperparameters fixed: $97{,}500$ adaptation steps (three epochs), TTT length $5$, and sequence length $4096$.

\subsubsection{Learning-rate robustness}
\label{sec:ablations:lr-sensitivity}

We found Osprey's initialization from a pretrained checkpoint may reduce sensitivity to adaptation hyperparameters, compared to random initialization like EAGLE3. Table~\ref{tab:lr-sensitivity} compares the warm-start Osprey drafter and the from-scratch EAGLE-3 baseline at two adaptation learning rates: $10^{-4}$, which is used for our main \texttt{Qwen3-8B} experiments, and $10^{-5}$. Reducing the learning rate by a factor of ten decreases the overall AL of the warm-start drafter by only $0.11$ ($3.686 \rightarrow 3.578$). In contrast, the from-scratch baseline loses $1.07$ AL ($3.176 \rightarrow 2.110$). These results suggest that the pretrained initialization provides a strong prior that can be refined during target-specific adaptation, making Osprey substantially more robust to the choice of learning rate.


\ifdefined\OspreyArxivStyle
\begin{table}[H]
\else
\begin{table}[t]
\fi
\centering
\caption{Learning-rate sensitivity on \texttt{Qwen3-8B}. 
Entries report mean accepted length (AL). When the adaptation learning rate 
drops from $10^{-4}$ to $10^{-5}$, the warm-start draft loses only $0.11$ AL, 
while the from-scratch baseline loses $1.07$ AL. Full per-domain 
acceptance-length matrices are reported in Appendix~\ref{app:lr-comparison}.}
\label{tab:lr-sensitivity}
\small
\setlength{\tabcolsep}{3.5pt}
\begin{tabular}{@{}llccc@{}}
\toprule
Draft & Metric & $10^{-5}$ & $10^{-4}$ & $\Delta$ \\
\midrule
\multirow{3}{*}{Ours}
 & ID  & 4.260 & 4.465 & $+0.21$ \\
 & OOD & 3.407 & 3.493 & $+0.09$ \\
 & All & 3.578 & 3.686 & $+0.11$ \\
\midrule
\multirow{3}{*}{Scratch}
 & ID  & 2.869 & 3.971 & $+1.10$ \\
 & OOD & 1.921 & 2.977 & $+1.06$ \\
 & All & 2.110 & 3.176 & $+1.07$ \\
\bottomrule
\end{tabular}
\end{table}

\subsubsection{How helpful is Stage~2 pretraining?}
\label{sec:ablations:pretrain-depth}

Osprey includes a target-agnostic pretraining stage (Stage~2) between layer pruning and target-specific adaptation. We investigate two questions: whether Stage~2 improves downstream performance and how much pretraining is needed to obtain most of its benefit.

\paragraph{Skipping target-agnostic pretraining entirely.}
Table~\ref{tab:pretrain-depth-ablation} isolates the contribution of Stage~2 by comparing a layer-pruned drafter that proceeds directly to target-specific adaptation and distillation (Stages~1, 3, and 4) with the complete pipeline that includes FineWeb pretraining (Stages~1--4). This ablation covers chat, code, finance, and math, yielding $16$ adaptation--evaluation domain pairs; commonsense is not included. Adding Stage~2 increases overall AL from $3.605$ to $3.790$ ($+0.185$). It improves both in-domain AL, from $4.306$ to $4.517$ ($+0.211$), and OOD AL, from $3.371$ to $3.547$ ($+0.176$). The improvement is consistent across all reported adaptation--evaluation domain pairs, indicating that target-agnostic pretraining provides a useful prior for subsequent target-specific adaptation.

\ifdefined\OspreyArxivStyle
\begin{table}[H]
\else
\begin{table}[t]
\fi
\centering
\caption{Effect of Stage~2 pretraining on Qwen3-8B accepted length (AL). Both variants use the same two-layer architecture and adaptation setup.}
\label{tab:pretrain-depth-ablation}
\ifdefined\OspreyArxivStyle
\scriptsize
\renewcommand{\arraystretch}{0.88}
\else
\small
\fi
\begin{tabular}{@{}llccc@{}}
\toprule
Train & Eval &
\shortstack{w/o Stage~2\\pretraining} &
\shortstack{w/ Stage~2\\pretraining} &
$\Delta$ \\
\midrule
\multirow{4}{*}{chat}
& chat        & 4.158 & \textbf{4.335} & $+0.177$ \\
& code        & 2.644 & \textbf{2.867} & $+0.223$ \\
& finance     & 3.935 & \textbf{4.124} & $+0.189$ \\
& math        & 3.888 & \textbf{4.057} & $+0.169$ \\
\midrule
\multirow{4}{*}{code}
& chat        & 3.735 & \textbf{3.844} & $+0.109$ \\
& code        & 3.804 & \textbf{4.145} & $+0.341$ \\
& finance     & 3.727 & \textbf{3.855} & $+0.128$ \\
& math        & 3.757 & \textbf{3.854} & $+0.097$ \\
\midrule
\multirow{4}{*}{finance}
& chat        & 4.091 & \textbf{4.235} & $+0.144$ \\
& code        & 2.964 & \textbf{3.096} & $+0.132$ \\
& finance     & 4.148 & \textbf{4.335} & $+0.187$ \\
& math        & 4.085 & \textbf{4.287} & $+0.202$ \\
\midrule
\multirow{4}{*}{math}
& chat        & 2.827 & \textbf{3.109} & $+0.282$ \\
& code        & 2.061 & \textbf{2.189} & $+0.128$ \\
& finance     & 2.737 & \textbf{3.048} & $+0.311$ \\
& math        & 5.113 & \textbf{5.253} & $+0.140$ \\
\midrule
\multicolumn{2}{@{}l}{ID Mean}
& 4.306 & \textbf{4.517} & $+0.211$ \\
\multicolumn{2}{@{}l}{OOD Mean}
& 3.371 & \textbf{3.547} & $+0.176$ \\
\multicolumn{2}{@{}l}{Overall}
& 3.605 & \textbf{3.790} & $+0.185$ \\
\bottomrule
\end{tabular}
\end{table}

\paragraph{Limited gains beyond 10k pretraining steps.}
Extending target-agnostic pretraining from $10$k to $55$k steps provides little additional benefit after target-specific adaptation. Under identical adaptation settings, the 10k-step checkpoint achieves an overall AL of $3.552$, compared with $3.578$ for the 55k-step checkpoint, a difference of only $0.026$. Their OOD-to-ID AL ratios are also nearly identical ($0.799$ versus $0.800$). Domain-level results and the corresponding pretraining curves are provided in Appendix~\ref{sec:appendix_pretraining}.

\subsubsection{Parameter-matched from-scratch baseline}
\label{sec:ablations:param-matched}

Our main experiments compare a two-layer warm-start Osprey drafter with a one-layer EAGLE-3 baseline, raising the possibility that Osprey's advantage results from its additional layer rather than from pretraining. Table~\ref{tab:param-matched-all-tps} separates these effects by including a parameter-matched two-layer EAGLE-3 drafter trained from scratch. As in the Stage~2 ablation, this comparison covers chat, code, finance, and math. All three drafters use the same adaptation data, optimizer, and learning rate.

The two-layer EAGLE-3 drafter achieves an overall AL of $3.385$, only $0.094$ higher than the one-layer EAGLE-3 drafter's $3.291$. Osprey achieves an overall AL of $3.790$, exceeding the parameter-matched two-layer EAGLE-3 drafter by $0.405$ despite using the same architecture. The corresponding OOD AL differences are $0.096$ between the one- and two-layer EAGLE-3 drafters and $0.408$ between Osprey and the two-layer EAGLE-3 drafter. These results indicate that Osprey's generalization advantage comes primarily from its pretrained language prior rather than its additional capacity.

\ifdefined\OspreyArxivStyle
\begin{table}[H]
\else
\begin{table}[!t]
\fi
\centering
\caption{Parameter-matched comparison on \texttt{Qwen3-8B}. Each cell reports AL/TPS.}
\label{tab:param-matched-all-tps}
\ifdefined\OspreyArxivStyle
\footnotesize
\renewcommand{\arraystretch}{0.94}
\else
\small
\fi
\begin{tabular}{@{}ll ccc@{}}
\toprule
Train & Eval & Eagle-3  & Eagle-3 & Ours\\
 &  &  1-layer &  2-layer & 2-layer \\
\midrule
\multirow{4}{*}{chat}
 & chat    & 3.755/380 & 3.661/328 & \textbf{4.335}/\textbf{394} \\
 & code    & 2.407/244 & 2.486/223 & \textbf{2.867}/\textbf{260} \\
 & fin& 3.558/365 & 3.542/332 & \textbf{4.124}/\textbf{380} \\
 & math    & 3.581/362 & 3.553/326 & \textbf{4.057}/\textbf{367} \\
\midrule
\multirow{4}{*}{code}
 & chat    & 3.399/345 & 3.541/331 & \textbf{3.844}/\textbf{349} \\
 & code    & 3.780/370 & 3.837/347 & \textbf{4.145}/\textbf{374} \\
 & fin& 3.469/355 & 3.558/333 & \textbf{3.855}/\textbf{354} \\
 & math    & 3.558/\textbf{370} & 3.633/345 & \textbf{3.854}/360 \\
\midrule
\multirow{4}{*}{finance}
 & chat    & 3.454/344 & 3.698/338 & \textbf{4.235}/\textbf{383} \\
 & code    & 2.516/253 & 2.614/240 & \textbf{3.096}/\textbf{277} \\
 & fin& 3.542/362 & 3.839/358 & \textbf{4.335}/\textbf{394} \\
 & math    & 3.736/375 & 3.825/354 & \textbf{4.287}/\textbf{383} \\
\midrule
\multirow{4}{*}{math}
 & chat    & 2.533/257 & 2.634/241 & \textbf{3.109}/\textbf{280} \\
 & code    & 1.857/188 & 2.014/180 & \textbf{2.189}/\textbf{193} \\
 & fin& 2.453/252 & 2.565/238 & \textbf{3.048}/\textbf{277} \\
 & math    & 5.058/\textbf{496} & 5.160/458 & \textbf{5.253}/463 \\
\midrule
\multicolumn{2}{@{}l}{ID Mean}
& 4.034/402 & 4.124/373 & \textbf{4.517}/\textbf{406} \\
\multicolumn{2}{@{}l}{OOD Mean}
& 3.043/309 & 3.139/290 & \textbf{3.547}/\textbf{322} \\
\multicolumn{2}{@{}l}{Overall}
& 3.291/332 & 3.385/311 & \textbf{3.790}/\textbf{343} \\
\bottomrule
\end{tabular}
\end{table}

\subsubsection{Computational justification and serving break-even}
\label{sec:ablations:cost}

We assess whether the upfront cost of target-agnostic pretraining can be justified through backbone reuse and serving-time compute savings. Stage~2 requires approximately $576$ H100-hours for $10$k pretraining steps and $3{,}040$ H100-hours for $55$k steps. Although it does not reduce the per-target adaptation budget, Stage~2 trains only the approximately $500$M-parameter shallow model, requires no target-model forward passes, and produces a backbone that can be reused across multiple target models (Table~\ref{tab:compute-accounting}). On \texttt{MiniMax-M2.5}, Osprey achieves a throughput of $219.8$ tokens/s, $17.5\%$ higher than EAGLE-3's $187.1$ tokens/s, saving approximately $880$ H100 GPU-hours per billion generated tokens. At this rate, the $55$k-step pretraining cost is recovered after approximately $3.5$ billion generated tokens, while the $10$k-step checkpoint breaks even after approximately $0.65$ billion tokens. Reusing the backbone across multiple targets can further reduce the effective break-even volume per target. However, the upfront cost may not be recovered in low-traffic or one-off deployments, as discussed in the Limitations.

\section{Conclusion}
State-of-the-art speculative drafters are typically trained for a single target model on narrow data distributions, requiring repeated training and limiting acceptance under domain shifts. We introduced \textbf{Osprey}, which prunes a pretrained small language model into a shallow backbone, restores its language-modeling capability through reusable target-agnostic pretraining, and then applies vocabulary alignment, zero-initialized QKV expansion, and target distillation. Compared with EAGLE-3, Osprey achieves mean AL that is $16.1\%$, $21.2\%$, and $22.7\%$ higher on Qwen3-8B, Llama-3.3-70B-Instruct, and MiniMax-M2.5, respectively, while delivering $17.5\%$ higher throughput on MiniMax-M2.5. Its strongest advantages appear on OOD and multilingual workloads, demonstrating that a broadly pretrained language prior provides an effective and reusable foundation for speculative drafters.

\section*{Limitations}

\textcolor{black}{We evaluate three targets---\texttt{Qwen3-8B}, \texttt{Llama-3.3-70B-Instruct}, and \texttt{MiniMax-M2.5}---under a fixed SGLang configuration; results may vary with other targets, hardware, batch sizes, or workloads.}
\textcolor{black}{Furthermore, the computational advantages of Osprey are conditional on inference workload: Stage~2 pretraining requires an upfront compute investment that is repaid only under continuous, high-volume serving or multi-target reuse. In low-traffic or one-off settings where total generated tokens are low, the upfront training compute is not amortized.}
The cross-domain study covers five domains and the MiniMax-M2.5 adaptation uses code-only target-generated data, which does not exhaust all languages, tasks, or deployment distributions. 
Most results are reported from single runs rather than multiple random seeds, so small differences should be interpreted cautiously. 
Because Osprey improves inference efficiency for existing LLMs, it may lower the cost of both beneficial and harmful uses and does not itself mitigate model bias, factuality, or safety issues.

\bibliography{custom}

\newpage
\appendix
\section*{Appendix}
\section{Detailed Related Work}
\label{sec:related-more}

\paragraph{Overview.}
This appendix expands the related-work paragraphs in Section~\ref{sec:related}. Briefly, Osprey's training recipe combines several established techniques: depth pruning and healing \citep{gromov2024unreasonable,men2024shortgpt,xia2024shearedllama}, vocabulary transfer across tokenizers \citep{minixhofer2022wechsel,dobler2023focus,fast_vocab_transfer,minixhofer2024zett,mundra2024vocabexp}, continued/over-budget pre-training \citep{hoffmann2022chinchilla,hu2024minicpm,allal2025smollm2,ibrahim2024continual,gururangan2020dapt}, and white-box/on-policy distillation \citep{hinton2015distilling,kim2016seqkd,gu2024minillm,agarwal2024gkd,zhou2024distillspec}. Drafter-free parallel decoding and multi-token prediction methods \citep{santilli2023parallel,fu2024lookahead,kou2024cllm,gloeckle2024mtp} are complementary since they modify the target or decoding loop, whereas Osprey improves the auxiliary drafter while leaving the target unchanged. The remainder of this appendix discusses each of these threads in detail.

\paragraph{Speculative decoding and target-conditioned drafters.}
Speculative decoding accelerates autoregressive inference by letting a cheaper draft model propose multiple future tokens that the target model verifies in parallel \citep{leviathan2023speculative,chen2023accelerating}.
Early approaches used specialized external draft models to propose candidate tokens for target-model verification \citep{xia2023specdec,miao2024specinfer}.
A second line of work conditions the drafter on the target's own hidden states: EAGLE \citep{li2024eagle} drafts at the penultimate-layer feature level, EAGLE-2 \citep{li2024eagle2} adds context-aware dynamic draft trees, and EAGLE-3 \citep{li2025eagle3} introduces multi-step training-time tests (TTT) together with a fusion of features tapped from three target layers.
Related architectures include the head-attached Medusa \citep{cai2024medusa} and its sequentially-dependent variant Hydra \citep{ankner2024hydra}, the RNN-based ReDrafter \citep{bhendawade2024redrafter} and Clover \citep{xiao2024clover,xiao2024clover2}, the parallel-drafting P-EAGLE \citep{hui2026peagle}, the semi-autoregressive Falcon \citep{gao2024falcon}, and feature-consistency HASS \citep{zhang2024hass}.
Osprey adopts EAGLE-3's TTT loss and three-layer hidden-state tap, but unlike all of the above we do not train the drafter body from scratch: the body inherits its parameters from a target-agnostic pre-trained next-token predictor, and target features are folded in via a zero-initialized expansion of the attention projections so that the pre-trained computation is preserved at the start of adaptation.

\paragraph{Self-speculative and target-internal drafters.}
A parallel line of work avoids an auxiliary drafter and draws proposals from the target itself.
LayerSkip \citep{elhoushi2024layerskip} trains the target so that its shallow layers form an accurate early exit, which then serves as the self-speculative drafter.
Draft \& Verify \citep{zhang2024draftverify} adaptively skips intermediate layers at inference time; SWIFT \citep{xia2024swift} chooses which layers to skip on-the-fly without any auxiliary training; and Kangaroo \citep{liu2024kangaroo} adds a small adapter on top of the target's shallow layers.
Speculative Streaming \citep{bhendawade2024specstreaming} fuses drafting into the target via multi-stream attention.
These designs eliminate the cost of training and serving a separate drafter, but they tightly couple drafter behavior to the target's parameters and tokenizer, so the same drafter cannot be reused across heterogeneous targets without retraining the target as well.
Osprey takes the opposite design choice---a standalone shallow drafter that is pre-trained once and lightly adapted per target.
We do borrow LayerSkip's observation that the first few transformer layers can be trained into competent next-token predictors; indeed, our Llama-based variant is built by depth-pruning a LayerSkip checkpoint and then pre-training it as an independent shallow LM.

\paragraph{Domain generalization and cross-target drafter reuse.}
A small but growing literature documents the failure mode our paper targets.
Spec-Bench \citep{sd-survey} reports large variance in accepted length across domains.
\citet{sandler2025disparate} formalize the resulting ``disparate impacts'' across user populations and propose a stochastic corrective fine-tuning fix.
\citet{sd-domain-shift} explicitly measure how single-domain draft fine-tuning degrades acceptance under domain shift.
Closest to our thesis, \citet{berdoz2025steering} argue that drafters initialized from a pre-trained small LM outperform from-scratch drafters under OOD inputs, and Mamba Drafters \citep{kim2025mamba} report similar gains for state-space drafters.
Two recent efforts target cross-target reuse from a different angle: OmniDraft \citep{chen2025omnidraft} maintains a single Llama-68M drafter across targets via online n-gram caching and cross-vocabulary verification, and \citet{timor2025universal} relax the shared-vocabulary assumption inside the verification step.
Online Speculative Decoding \citep{liu2024osd} continuously fine-tunes the drafter from observed traffic.
Osprey differs from this body of work along two axes.
First, the broad prior is acquired by an explicit large-scale pre-training stage on generic web text \citep{penedo2024fineweb}, rather than by online updates, off-the-shelf SLM borrowing, or selection among multiple pre-trained drafters.
Second, target-feature ingestion is added via a zero-initialized QKV expansion that keeps the drafter functionally identical to its pre-trained checkpoint at the start of adaptation, which lets us isolate the contribution of pre-training from the contribution of target conditioning and reuse the same pre-trained checkpoint across an arbitrary number of targets.

\paragraph{Knowledge distillation for autoregressive drafters.}
Our adaptation stage is structurally a knowledge-distillation problem: aligning the drafter's output distribution to the frozen target's at each predicted position.
The general framework is white-box distillation \citep{hinton2015distilling,kim2016seqkd}, and modern autoregressive variants close the train/inference distribution gap with on-policy data and reverse KL \citep{gu2024minillm,agarwal2024gkd}.
The most direct precedent for drafter training is DistillSpec \citep{zhou2024distillspec}, which uses on-policy white-box distillation to improve speculative acceptance and reports favorable generalization to unseen tasks.
Osprey's Stage 4 inherits the EAGLE-3 TTT loss, which can be viewed as an on-policy multi-step distillation against the target.
Our contribution is therefore not a new distillation objective; it is a stronger initialization for the student.
By starting from a pre-trained shallow LM rather than from random weights, the same distillation budget yields drafters that preserve substantially more linguistic behaviour outside the distillation distribution.

\paragraph{Compact pre-training: depth pruning, vocabulary transfer, and over-budget training.}
Stages 1--3 of Osprey assemble three established sub-fields into one pipeline.
\emph{Depth pruning.} \citet{gromov2024unreasonable} and \citet{men2024shortgpt} show that a large fraction of an LLM's layers can be removed with only modest degradation after light healing, and \citet{xia2024shearedllama} extend this to a prune-then-continued-pretrain pipeline for sub-billion Llamas.
Our Stage 1 keeps only the first $N\in\{2,4\}$ transformer blocks, in line with this evidence and with LayerSkip's early-exit-friendly initialization.
\emph{Vocabulary transfer.} When the source draft's tokenizer differs from the target's, we initialize new embedding rows from existing draft rows, in the lineage of WECHSEL \citep{minixhofer2022wechsel}, FOCUS \citep{dobler2023focus}, FVT \citep{fast_vocab_transfer}, and the zero-shot tokenizer transfer of \citet{minixhofer2024zett}.
Our average-of-subtokens initialization is in the spirit of FVT and is competitive with more elaborate alternatives \citep{mundra2024vocabexp} while requiring no auxiliary corpus.
\emph{Continued and over-budget pre-training.} Stage 2 is large-scale continued pre-training on FineWeb \citep{penedo2024fineweb}, deliberately trained past the compute-optimal regime of \citet{hoffmann2022chinchilla}---a regime that other recent sub-2B models have also found beneficial \citep{hu2024minicpm,allal2025smollm2}.
Practical recipes for continued pre-training \citep{ibrahim2024continual,gururangan2020dapt} inform our learning-rate and replay choices.
We are not aware of prior work that combines these three steps into a draft-model recipe for speculative decoding.

\paragraph{Drafter-free parallel decoding.}
A complementary class of methods accelerates inference without an auxiliary drafter at all.
Jacobi-style parallel decoding \citep{santilli2023parallel,fu2024lookahead} and its distilled fixed-point variants \citep{kou2024cllm} replace token-by-token decoding with parallel updates, and multi-token-prediction approaches \citep{gloeckle2024mtp} train the target itself with auxiliary heads that emit several future tokens at once.
These methods are orthogonal to Osprey: they modify the target or the decoding loop, whereas Osprey leaves the target untouched and improves the drafter.

\section{Implementation Notes}

Our codebase implements the method in SpecForge on top of SGLang.
The EAGLE-3 training path uses online target hidden-state generation, FSDP draft training, and optional tensor parallelism for the target model.
The conversion scripts preserve pretrained layers, expand Q/K/V projections for the EAGLE-3 dual-stream interface, and optionally swap tokenizer rows into the target vocabulary.
The SGLang evaluation path launches the target model with EAGLE-3 speculative decoding and records live accepted length plus throughput.
For the ablation matrix, all architectures share the same target, training budget, training domains, evaluation domains, and speculative serving configuration.

\section{Pretraining Implementation Details}
\label{sec:appendix_pretraining}

This appendix documents the hardware, optimization, data pipeline, and compute budget for the draft-pretraining stage.
All configuration values are taken verbatim from the released training configs and \texttt{slurm/train.slurm.sh}.

\paragraph{Hardware and precision.}
Each pretraining run uses \textbf{4 nodes $\times$ 8 NVIDIA H100 GPUs = 32 GPUs} (world size 32), launched through Slurm with one task per node and \texttt{torchrun} controlling intra-node ranks.
We train data-parallel under PyTorch FSDP with sharded optimizer states and gradients to fit the optimizer footprint on H100 80GB.
All compute runs in \texttt{bfloat16} with TensorFloat-32 matmuls enabled.
Attention is computed by FlashAttention-2 \citep{dao2023flashattention2}, which transparently dispatches to the Hopper-specific FlashAttention-3 kernel on H100.

\paragraph{Optimizer and schedule.}
We optimize with AdamW \citep{loshchilov2019adamw} using
$(\beta_1, \beta_2, \epsilon) = (0.9,\, 0.95,\, 10^{-8})$,
weight decay $0.1$, and gradient-norm clipping at $1.0$.
The learning rate follows a cosine schedule with a $1\%$ linear warmup ($600$ of $60{,}000$ steps) and peak rate $1{\times}10^{-4}$.
After exploratory experiments with constant-then-decay schedules, we converged on plain cosine for stability across both the LayerSkip-Llama and Qwen3 drafts.

\paragraph{Batching and token budget.}
Sequence length is fixed at $4096$ tokens, which covers the dominant document length in FineWeb while keeping attention cost manageable at 32-GPU scale.
The per-device micro-batch is $4$, with gradient accumulation $16$, giving an effective global batch of
\begin{multline*}
  32 \times 4 \times 16 \times 4096 \;\approx\; 8.39\,\text{M tokens}
\end{multline*}  
for every optimizer step.
Training runs for $60{,}000$ optimizer steps, corresponding to $\approx 503$B tokens (about five passes over FineWeb sample-100BT). Each pretraining run takes approximately $80$--$110$ wall-clock hours on the 32-GPU configuration.

\paragraph{Data pipeline.}
The FineWeb sample-100BT \citep{penedo2024fineweb} corpus is tokenized once with the \emph{draft's own} tokenizer (Llama-3.2 for the LayerSkip-Llama draft, Qwen3-4B for the Qwen draft) and serialized into per-shard \texttt{uint32} memory-mapped binaries with documents separated by a single EOS token.
A multi-shard, infinite random-window sampler exposes fixed-length $4096$-token training examples to the data loader, so different ranks see disjoint window offsets within and across shards.
This format avoids re-tokenization on the training path and lets us re-use the same on-disk corpus across both draft families.

\paragraph{Layer-pruning configurations.}
We instantiate two pruned drafts.
On \texttt{facebook/layerskip-llama3.2-1B} \citep{elhoushi2024layerskip} we keep $N{=}4$ of $16$ blocks ($\approx$506M parameters);
on \texttt{Qwen/Qwen3-4B} \citep{qwen3} we keep $N{=}2$ of $36$ blocks ($\approx$591M parameters).
For both drafts we inherit embeddings, the final RMSNorm, and the LM head verbatim, patch \texttt{num\_hidden\_layers} to $N$, and truncate architecture-specific per-layer metadata (e.g., Qwen3's \texttt{layer\_types}) accordingly.
The LayerSkip variant is chosen deliberately: at step~0 the four-layer pruned model has a substantially stronger local next-token prior than a standard Llama-3.2-1B truncation, which we observe directly in the initial training loss ($\sim$4.5 vs.\ $\sim$13.5 nats).

\ifdefined\OspreyArxivStyle
\begin{figure}[H]
\centering
\begin{minipage}[t]{0.485\textwidth}
\centering
\includegraphics[width=\linewidth]{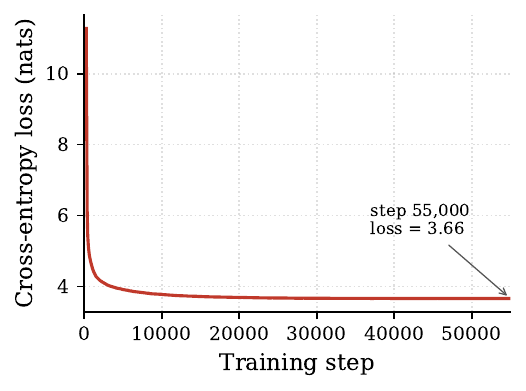}
\captionof{figure}{Pretraining cross-entropy loss on FineWeb. Qwen3-4B 2L, 55k steps.}
\label{fig:train-loss}
\end{minipage}\hfill
\begin{minipage}[t]{0.485\textwidth}
\centering
\includegraphics[width=\linewidth]{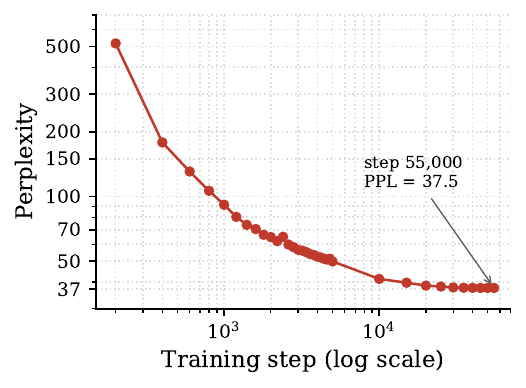}
\captionof{figure}{Held-out perplexity during pretraining. PPL plateaus near $37$ past step $10{,}000$.}
\label{fig:ppl}
\end{minipage}
\end{figure}
\else
\begin{figure}[t]
\centering
\includegraphics[width=0.8\columnwidth]{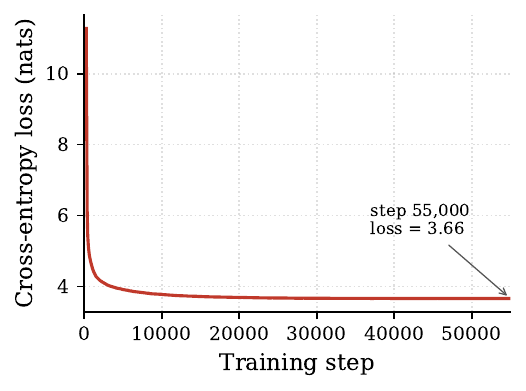}
\caption{Pretraining cross-entropy loss on FineWeb.
\textcolor[HTML]{c0392b}{Red}: Qwen3-4B 2L (55k steps).}
\label{fig:train-loss}
\centering
\includegraphics[width=0.8\columnwidth]{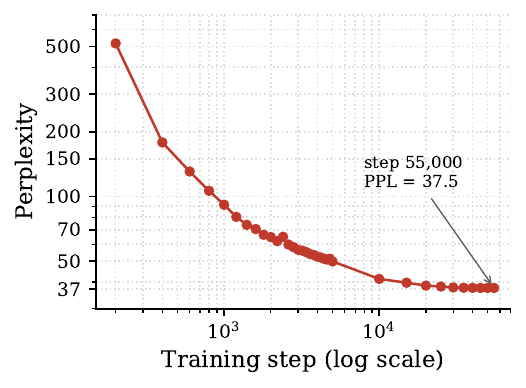}
\caption{Held-out perplexity of the Qwen3-4B 2-layer drafter during pretraining (log-scale axes).
PPL falls from $\sim$520 to a plateau near $37$ past step $10{,}000$.}
\label{fig:ppl}
\end{figure}
\fi

\subsection{Pretraining-budget saturation}
Our full pretraining run uses $55$k steps ($\approx462$B tokens), but Table~\ref{tab:qwen3-2layer-10k-55k} shows that a $10$k-step checkpoint already achieves nearly the same downstream performance as the $55$k-step checkpoint. Extending pretraining from $10$k to $55$k steps improves overall AL by only $0.026$, while OOD/ID retention remains effectively unchanged at approximately $0.80$.

\ifdefined\OspreyArxivStyle
\begin{table}[H]
\else
\begin{table}[t]
\fi
\centering
\caption{Effect of the pretraining budget. The 10k-step and 55k-step checkpoints are adapted identically to \texttt{Qwen3-8B} using $\text{lr}{=}10^{-5}$. The additional 45k pretraining steps improve overall AL by only $0.026$. Domain rows report AL for the corresponding adaptation split, averaged across evaluation domains.}
\label{tab:qwen3-2layer-10k-55k}
\small
\setlength{\tabcolsep}{5pt}
\begin{tabular}{@{}lccc@{}}
\toprule
Adaptation split / metric & 10k ckpt & 55k ckpt & $\Delta$ \\
\midrule
chat        & 3.619 & \textbf{3.639} & $+0.020$ \\
code        & 3.684 & \textbf{3.714} & $+0.030$ \\
commonsense & 3.271 & \textbf{3.316} & $+0.045$ \\
finance     & 3.760 & \textbf{3.779} & $+0.019$ \\
math        & 3.424 & \textbf{3.440} & $+0.016$ \\
\midrule
Overall AL  & 3.552 & \textbf{3.578} & $+0.026$ \\
ID AL       & 4.230 & \textbf{4.260} & $+0.030$ \\
OOD AL      & 3.382 & \textbf{3.407} & $+0.025$ \\
OOD/ID ret. & 0.799 & \textbf{0.800} & $+0.000$ \\
\bottomrule
\end{tabular}
\end{table}

The upstream pretraining curves provide further evidence of this saturation. Figure~\ref{fig:train-loss} shows that the cross-entropy loss drops sharply from approximately $19$ to $5$ nats during the first $1$k steps and then decreases more gradually to approximately $3.7$ nats. Figure~\ref{fig:ppl} shows that held-out perplexity plateaus near $37$ after approximately $10{,}000$ steps. Together with the downstream results, these curves indicate that a relatively short pretraining run captures most of the benefit, making the cost of producing a reusable shallow NTP backbone modest relative to its downstream gains.


\section{Adaptation Implementation Details}
\label{app:compute} 

All experiments use bf16 mixed-precision training under PyTorch FSDP and
SGLang serving with EAGLE-3 speculative decoding. Hardware differs by
target model.

\paragraph{Qwen3-8B target (train and inference).}
Stages~3--4 (per-target
adaptation and on-policy distillation against Qwen3-8B) use
\textbf{2~NVIDIA~H100~80\,GB GPUs} on a single node, with the target
served in bf16 via SGLang to produce on-policy hidden states and
logits. Evaluation (acceptance length and decoding throughput in
Sec.~\ref{sec:experiments}) is also run on a \textbf{single H100~80\,GB
GPU} (\texttt{tp=1}) under SGLang, matching the canonical EAGLE-3
serving configuration.

\paragraph{MiniMax-M2.5 target (train and inference).}
For the 229B-parameter MoE target, both adaptation/distillation and
evaluation are performed on \textbf{4~NVIDIA~B200 GPUs} on a single
node. The MiniMax-M2.5 target is served in FP8 under SGLang with
\texttt{tensor\_parallel\_size=4} and \texttt{expert\_parallel\_size=4},
fitting the model into the 4$\times$192\,GB HBM3e of the B200 node. The
draft trains on the same 4~B200 GPUs against MiniMax-generated code data for 35k steps. We use SGLang's FP8 KV-cache path
(\texttt{sglang\_fp4kv} environment) to keep target memory headroom for
on-policy rollouts during distillation.

\begingroup
\color{black}
\noindent\begin{minipage}{\columnwidth}
\paragraph{Per-target adaptation cost.}
For each Qwen3-8B domain, Osprey and the matched scratch EAGLE-3
baseline use identical data, steps, and hardware, with near-matched
drafter bodies (265\,M vs.\ 269\,M).

\begin{center}
\color{black}
\scriptsize
\setlength{\tabcolsep}{3pt}
\renewcommand{\arraystretch}{1.05}
\captionsetup{labelfont={color=black},textfont={color=black}}
\captionof{table}{Per-domain Qwen3-8B adaptation workload.}
\label{tab:adaptation-cost}
\begin{tabular}{@{}p{0.32\columnwidth}p{0.60\columnwidth}@{}}
\toprule
Quantity & Value \\
\midrule
Unique prompts & 65k; responses regenerated by the target \\
Sequence / TTT length & 4096 / 5 \\
Optimizer steps & 97.5k (3 epochs) \\
Tokens processed & $\approx$0.8B \\
Target forward passes & $\approx$195k; hidden states and logits computed online \\
Hardware & 2$\times$H100; target co-served in bf16 \\
\bottomrule
\end{tabular}
\end{center}

For Llama-3.3-70B, Osprey and the scratch baseline likewise use the
same target-regenerated data and 37.5k steps. MiniMax-M2.5 uses 70k
target-regenerated code prompts and 35k steps on 4$\times$B200; because
the public comparator's training cost is not reported, we make no
cost-parity claim for that comparison.
\end{minipage}
\endgroup

\paragraph{Serving configuration.}
All reported acceptance lengths and throughputs use
\texttt{batch\_size=1}, \texttt{steps=5}, \texttt{topk=1},
\texttt{num\_draft\_tokens=6}, and \texttt{max\_tokens=4096} in
SGLang. The Qwen3-8B serving configuration is \texttt{tp=1} on one
H100; the MiniMax-M2.5 configuration is \texttt{tp=4, ep=4} on
4~B200. The same draft checkpoint is reused across the train and
evaluation phases without further fine-tuning at inference time.

\ifdefined\OspreyArxivStyle
  \clearpage
\fi
\begingroup
\color{black}
\begin{samepage}
\subsection{Serving-Configuration Sensitivity}
\label{app:serving-sensitivity}

We vary one serving axis at a time on Qwen3-8B for Osprey and the EAGLE-3 baseline. Because SGLang couples chain steps and
the number of draft tokens when \texttt{topk=1}, we sweep these two
values together.

\begin{center}
\centering
\color{black}
\scriptsize
\setlength{\tabcolsep}{3pt}
\renewcommand{\arraystretch}{1.05}
\captionsetup{labelfont={color=black},textfont={color=black}}
\captionof{table}{Serving sensitivity on one H100. Each cell reports accepted
length (AL) / decoding throughput (tok/s). Unspecified settings use
batch size 1, 5 steps, 6 draft tokens, temperature 0, and top-$k$ 1.}
\label{tab:serving-sensitivity}
\begin{tabular}{@{}llcc@{}}
\toprule
Axis & Setting & \shortstack{EAGLE-3\\AL / tok/s} &
\shortstack{Osprey\\AL / tok/s} \\
\midrule
\multirow{4}{*}{Batch size}
  & 1  & 2.34 / 217  & 3.29 / 270  \\
  & 4  & 2.36 / 785  & 3.32 / 970  \\
  & 8  & 2.35 / 1454 & 3.34 / 1716 \\
  & 16 & 2.35 / 2534 & 3.28 / 2855 \\
\midrule
\multirow{4}{*}{\shortstack[l]{Steps,\\draft tokens}}
  & $(1,2)$ & 1.67 / 182 & 1.79 / 187 \\
  & $(3,4)$ & 2.22 / 227 & 2.76 / 262 \\
  & $(5,6)$ & 2.34 / 217 & 3.29 / 270 \\
  & $(7,8)$ & 2.36 / 202 & 3.61 / 258 \\
\midrule
\multirow{3}{*}{Temperature}
  & 0   & 2.34 / 217 & 3.29 / 270 \\
  & 0.7 & 2.29 / 207 & 3.21 / 258 \\
  & 1.0 & 2.17 / 195 & 2.92 / 231 \\
\bottomrule
\end{tabular}
\end{center}

Accepted length is stable across batch sizes, while Osprey's throughput
gain remains positive (24\% at batch size 1 and 13\% at batch size 16).
Increasing the chain from 2 to 8 draft tokens widens Osprey's accepted-
length gain from 7\% to 53\%, as EAGLE-3 saturates near 6 tokens. The
gain also persists under sampling (41\% at temperature 0 and 35\% at
temperature 1.0).
\end{samepage}
\endgroup

\paragraph{\textcolor{black}{Compute and Resource Accounting.}}
\textcolor{black}{Table~\ref{tab:compute-accounting} provides the complete accounting of accelerator-hours, wall-clock time, tokens/steps, and target forward passes across pretraining and per-target adaptation stages. While Stage~2 pretraining requires an upfront investment, per-target distillation incurs the same training cost as standard EAGLE-3 training.}

\textcolor{black}{\textbf{Why Not Scale From-Scratch Distillation?} Expanding the training budget of a target-conditioned drafter to hundreds of billions of tokens is compute-prohibitive due to target forward pass overhead. Specifically, per-token FLOPs during target distillation are dominated by running the target model: $\text{FLOPs/token} \approx 2(N_{\target} + N_{\draft})$. For large targets (e.g., hundreds of billions of parameters such as MiniMax-M2.5 at 229B), distilling over hundreds of billions of tokens would require hundreds of thousands of GPU-hours. In contrast, Osprey's Stage~2 pretraining operates solely on the shallow drafter ($\text{FLOPs/token} \approx 6 N_{\draft}$), requiring only $\approx 3{,}040$ H100-hours for 462B tokens with zero target forward passes. Target-agnostic pretraining is therefore orders of magnitude more compute-efficient for instilling general language priors than on-target distillation, and this learned prior is subsequently shared across multiple target architectures.}

\begin{table*}[t]
\centering
\small
\setlength{\tabcolsep}{4pt}
\resizebox{\textwidth}{!}{%
\begin{tabular}{@{}lccccc@{}}
\toprule
\textcolor{black}{\textbf{Stage / Target}} & \textcolor{black}{\textbf{Hardware}} & \textcolor{black}{\textbf{Wall-clock}} & \textcolor{black}{\textbf{GPU-Hours}} & \textcolor{black}{\textbf{Tokens / Steps}} & \textcolor{black}{\textbf{Target Fwd Passes}} \\
\midrule
\textcolor{black}{\textbf{Stage 2 Pretraining (55k)}} & \textcolor{black}{32$\times$ H100} & \textcolor{black}{$\approx 95$ h} & \textcolor{black}{3,040 h} & \textcolor{black}{462B tokens} & \textcolor{black}{0} \\
\textcolor{black}{\textbf{Stage 2 Pretraining (10k)}} & \textcolor{black}{32$\times$ H100} & \textcolor{black}{$\approx 18$ h} & \textcolor{black}{576 h} & \textcolor{black}{84B tokens} & \textcolor{black}{0} \\
\midrule
\textcolor{black}{\textbf{Qwen3-8B Adaptation (Osprey)}} & \textcolor{black}{2$\times$ H100} & \textcolor{black}{$\approx 24$ h} & \textcolor{black}{48 h} & \textcolor{black}{97.5k steps} & \textcolor{black}{97.5k rollouts} \\
\textcolor{black}{\textbf{Qwen3-8B Scratch (EAGLE-3)}} & \textcolor{black}{2$\times$ H100} & \textcolor{black}{$\approx 24$ h} & \textcolor{black}{48 h} & \textcolor{black}{97.5k steps} & \textcolor{black}{97.5k rollouts} \\
\midrule
\textcolor{black}{\textbf{MiniMax-M2.5 Adaptation (Osprey)}} & \textcolor{black}{4$\times$ B200} & \textcolor{black}{$\approx 36$ h} & \textcolor{black}{144 h} & \textcolor{black}{35k steps} & \textcolor{black}{35k rollouts} \\
\textcolor{black}{\textbf{MiniMax-M2.5 Scratch (EAGLE-3)}} & \textcolor{black}{4$\times$ B200} & \textcolor{black}{$\approx 36$ h} & \textcolor{black}{144 h} & \textcolor{black}{35k steps} & \textcolor{black}{35k rollouts} \\
\bottomrule
\end{tabular}
}
\caption{\textcolor{black}{End-to-end compute and resource accounting for Osprey pretraining and target adaptation compared with from-scratch EAGLE-3 baselines.}}
\label{tab:compute-accounting}
\end{table*}

\section{Draft Model Sizes}
\label{app:draft-sizes}
We report body parameters (transformer layers + 3-layer hidden
combiner $f_c$), excluding the input embedding and LM head.
\begin{table}[h]
\centering
\footnotesize
\setlength{\tabcolsep}{4pt}
\renewcommand{\arraystretch}{1.05}
\begin{tabular}{@{}lccr@{}}
\toprule
Draft & $L$ & $d_{\text{h}}$ & Body \\
\midrule
EAGLE-3 1L         & 1 & 4096 & 269\,M \\
EAGLE-3 2L         & 2 & 4096 & 487\,M \\
\textbf{Osprey}    & 2 & 2560 & \textbf{265\,M} \\
LayerSkip          & 4 & 2048 & 294\,M \\
\bottomrule
\end{tabular}
\caption{Draft sizes on Qwen3-8B (body+$f_c$ only; embedding and
LM head excluded for fair comparison). $L$ = layers, $d_h$ = hidden.}
\label{tab:draft-sizes}
\end{table}
Under this matched count, Osprey 2L (265\,M) is essentially the same
size as the 1-layer EAGLE-3 scratch baseline (269\,M) and
\emph{smaller} than the 2-layer parameter-matched scratch baseline
(487\,M).



\section{Learning-Rate Comparison Matrices}
\label{app:lr-comparison}

For comparison with Table~\ref{tab:lr-sensitivity}, we report the full
per-domain acceptance-length matrices in Table~\ref{tab:lr-full-acceptance}
for the two drafts used in the learning-rate sensitivity study.
Each row is one single-domain adapted drafter evaluated on all five domains.
Cells report acceptance length only; throughput is omitted for compactness.

\begin{table*}[t]
\centering
\caption{Full acceptance-length matrices on \texttt{Qwen3-8B} for
$\mathrm{lr}=10^{-5}$ and $\mathrm{lr}=10^{-4}$.  \textit{Warm-start}
denotes \texttt{qwen3\_2layer}; \textit{Scratch} denotes
\texttt{vanilla\_1layer}. Bold marks the in-domain train$=$eval cell.}
\label{tab:lr-full-acceptance}
\scriptsize
\setlength{\tabcolsep}{3.2pt}
\renewcommand{\arraystretch}{1.08}
\resizebox{\textwidth}{!}{%
\begin{tabular}{@{}lllccccccc@{}}
\toprule
LR & Draft & Train & Chat & Code & Comm. & Finance & Math & Mean & OOD \\
\midrule
\multirow{10}{*}{$10^{-5}$}
 & \multirow{5}{*}{Warm-start}
 & chat        & \textbf{4.171} & 2.688 & 3.729 & 3.876 & 3.731 & 3.639 & 3.506 \\
 & & code        & 3.840 & \textbf{3.822} & 3.571 & 3.717 & 3.621 & 3.714 & 3.687 \\
 & & commonsense & 3.795 & 2.222 & \textbf{4.116} & 3.420 & 3.029 & 3.316 & 3.116 \\
 & & finance     & 4.141 & 2.911 & 3.756 & \textbf{4.119} & 3.965 & 3.779 & 3.693 \\
 & & math        & 3.376 & 2.170 & 3.415 & 3.166 & \textbf{5.073} & 3.440 & 3.032 \\
\cmidrule(lr){2-10}
 & \multirow{5}{*}{Scratch}
 & chat        & \textbf{1.621} & 1.446 & 1.602 & 1.604 & 1.734 & 1.601 & 1.597 \\
 & & code        & 2.245 & \textbf{3.097} & 2.278 & 2.411 & 2.638 & 2.534 & 2.393 \\
 & & commonsense & 2.137 & 1.441 & \textbf{2.625} & 1.960 & 1.525 & 1.938 & 1.766 \\
 & & finance     & 1.939 & 1.814 & 1.976 & \textbf{2.136} & 2.393 & 2.052 & 2.031 \\
 & & math        & 1.733 & 1.631 & 2.103 & 1.809 & \textbf{4.864} & 2.428 & 1.819 \\
\midrule
\multirow{10}{*}{$10^{-4}$}
 & \multirow{5}{*}{Warm-start}
 & chat        & \textbf{4.335} & 2.867 & 3.867 & 4.124 & 4.057 & 3.850 & 3.729 \\
 & & code        & 3.844 & \textbf{4.145} & 3.538 & 3.855 & 3.854 & 3.847 & 3.773 \\
 & & commonsense & 3.793 & 2.246 & \textbf{4.259} & 3.477 & 3.228 & 3.401 & 3.186 \\
 & & finance     & 4.235 & 3.096 & 3.876 & \textbf{4.335} & 4.287 & 3.966 & 3.874 \\
 & & math        & 3.109 & 2.189 & 3.266 & 3.048 & \textbf{5.253} & 3.373 & 2.903 \\
\cmidrule(lr){2-10}
 & \multirow{5}{*}{Scratch}
 & chat        & \textbf{3.755} & 2.407 & 3.245 & 3.558 & 3.581 & 3.309 & 3.198 \\
 & & code        & 3.399 & \textbf{3.780} & 3.182 & 3.469 & 3.558 & 3.478 & 3.402 \\
 & & commonsense & 3.268 & 1.951 & \textbf{3.722} & 2.997 & 2.543 & 2.896 & 2.690 \\
 & & finance     & 3.454 & 2.516 & 3.113 & \textbf{3.542} & 3.736 & 3.272 & 3.205 \\
 & & math        & 2.533 & 1.857 & 2.721 & 2.453 & \textbf{5.058} & 2.924 & 2.391 \\
\bottomrule
\end{tabular}%
}
\end{table*}
\section{Cross-Domain Warm-Start Ablation under DFlash}
\label{app:dflash}

Diffusion draft models have recently emerged as a promising alternative to
autoregressive drafters in speculative decoding, motivated by the strong
parallel-generation behaviour of diffusion language
models~\citep{chen2026dflash}. The question of speculative generalisation is
therefore not specific to autoregressive drafters. This appendix asks whether
a target-agnostic warm start also helps under a diffusion-style drafting
interface.

The ablation first contrasts two initialisations of the same drafter under the
original DFlash-style interface: a random-body baseline and a warm start from
a diffusion language model. We then introduce an EAGLE-style DFlash variant
that preserves the pretrained drafter attention pathway at initialization, in
order to test whether reducing the target-feature interface mismatch improves
cross-domain transfer.

\subsection{Background on DFlash}
\label{app:dflash:method}

DFlash trains a block-parallel drafter whose self-attention ingests target
hidden states through a key/value injection path rather than EAGLE-3's stream-swap channel \cite{chen2026dflash}. At each block of size $B$, the drafter consumes one
verified anchor token concatenated with $B{-}1$ \texttt{[MASK]} placeholders
and emits $B{-}1$ proposal tokens per forward pass, with target features
broadcast into every block. The proposal distribution at each masked position
is therefore conditioned on the verified anchor, the local target features
and, through the intra-block self-attention, the surrounding masked positions.

\subsection{IDLM as a Diffusion Warm-Start Source}
\label{app:dflash:idlm}

A warm-start source for DFlash must itself be a diffusion language model,
since DFlash drafts a block of tokens in parallel rather than
autoregressively. We adopt the Introspective Diffusion Language Model
(IDLM) of~\citet{yu2026introspective}, a recent diffusion language model
with strong small-model performance.

This diffusion warm-start source is analogous to the Shallow NTP used by
Osprey in that it is target-agnostic, but it is not an NTP checkpoint: it is
trained with a diffusion language-modeling objective and is used only for the
DFlash ablation.

IDLM is decoded via \emph{introspective strided decoding} (ISD), in which the
model unifies decoding and verification of mask tokens in the same forward
step. This decoding pattern is structurally similar to speculative decoding,
making IDLM a natural warm-start source for a block-parallel drafter.

The publicly released IDLM checkpoints are too large to use directly as draft
models for speculative decoding. We instead train a compact diffusion language
model by applying the IDLM training recipe to a four-layer LayerSkip-Llama
backbone~\citep{elhoushi2024layerskip}. Training uses NVIDIA's
Nemotron-Post-Training-Dataset-v2~\citep{NemotronPostTrainingDatasetV2,nvidia2025nvidianemotronnano2}.
This checkpoint is used as the warm-start source from which the DFlash
drafter body weights are loaded; the random-body baseline shares everything
except this body initialization.

\subsection{Setup}
\label{app:dflash:setup}

In this ablation, ``body'' refers only to the transformer blocks, excluding
the input embedding and LM head.

\paragraph{Original DFlash-style configurations.}
\begin{itemize}
  \item \textbf{Baseline (random body).} Four-layer Llama-shape drafter
        with a randomly initialized transformer-block body. The embedding
        and LM head are initialized from the same IDLM source as in the
        warm-start configuration and row-mapped to the Qwen3 vocabulary, so
        the ablation isolates the effect of body initialization.

  \item \textbf{Warm start (IDLM).} Identical to the baseline except that
        the transformer-block body is initialized from the compact IDLM-style
        diffusion checkpoint.
\end{itemize}

The two configurations share architecture, embedding initialization,
LM-head initialization, training data and optimizer; the only variable is
whether the transformer-block body starts from IDLM-style diffusion weights
or from random weights.

\paragraph{Target, training and evaluation.}
The target is Qwen/Qwen3-8B. Each configuration is fine-tuned for six epochs
on the same 65k single-domain math split used in Section~3 (Qwen3 thinking
chat template, sequence length $4096$, effective batch $32$, peak learning
rate $6{\times}10^{-4}$ with cosine decay to $6{\times}10^{-5}$ and warmup
ratio $0.04$, block size $B{=}16$). Cross-domain accepted length is
reported on the five-domain evaluation protocol of the main paper
(\emph{math, chat, code, commonsense, finance}), with $512$ prompts per
domain through live SGLang.

\subsection{EAGLE-Style Target Interface for DFlash}
\label{app:dflash:eagle}

The original DFlash-style interface feeds target-model features into the
drafter through target-derived key/value streams. This creates a mismatch for
a pretrained IDLM body: during diffusion pretraining, its attention layers see
key/value representations produced by the drafter itself, whereas during
DFlash fine-tuning they receive key/value inputs derived from the target
model. This key/value distribution drift can reduce the effective reuse of
pretrained attention computation, limiting the gain from body warm-starting.

To test this hypothesis, we add an EAGLE-style DFlash variant that keeps the
baseline DFlash path untouched but changes how target features enter each
attention layer. Each attention layer keeps the pretrained drafter projections
and adds zero-initialized target-side projections:
\[
\begin{aligned}
Q &= W_Q h + W_Q^{\mathrm{tgt}} c_t, \\
K &= W_K h + W_K^{\mathrm{tgt}} c_t, \\
V &= W_V h + W_V^{\mathrm{tgt}} c_t .
\end{aligned}
\]
where $h$ is the drafter hidden state and $c_t$ is the projected target
feature. At initialization, the target-side projections contribute nothing,
so the attention computation is identical to a pretrained-only forward.
During fine-tuning, the zero channel gradually learns to incorporate target
features.

The target feature $c_t$ is gathered once at the block's verified anchor
position and broadcast across the block. This avoids inserting extra
target-derived key/value positions into intra-block attention; the block
attention remains the drafter's own self-attention path.

We also match the IDLM pretraining convention by using causal intra-block
attention and shift-by-one labels: position $k$ predicts the token at
$k{+}1$. This contrasts with the original DFlash setting, which uses
bidirectional block attention and same-position labels. Both changes preserve
the pretrained attention and label conventions more closely than the original
DFlash-style interface.

\subsection{Results}
\label{app:dflash:results}

Table~\ref{tab:dflash-controlled} reports all DFlash ablations. The first two
columns compare random-body and IDLM-warm-start initialization under the
original DFlash-style interface. The last column reports the EAGLE-style
DFlash variant under the aligned HF protocol.

\begin{table}[t]
  \centering
  \small
  \setlength{\tabcolsep}{5pt}
  \begin{tabular}{lccc}
    \toprule
    Eval 
      & \multicolumn{2}{c}{Original DFlash-style}
      & EAGLE-style DFlash \\
    \cmidrule(lr){2-3}
     domain & Baseline & IDLM WS & IDLM WS \\
    \midrule
    \textbf{math} (ID) & 5.96 & \textbf{6.03} & 5.25 \\
    chat        & 1.50 & 1.53 & \textbf{1.88 }\\
    code        & 1.58 & 1.63 & \textbf{1.93} \\
    common & 1.88 & 1.92 & \textbf{2.21} \\
    finance     & 1.60 & 1.65 & \textbf{1.99 }\\
    \midrule
    OOD avg. & 1.64 & 1.68 & \textbf{2.00} \\
    \bottomrule
  \end{tabular}
  \caption{Cross-domain accepted length under DFlash variants. The original
    DFlash-style columns compare a random-body baseline with an IDLM WS (warm
    start); the two configurations share architecture, embedding
    initialization, LM-head initialization, training data and optimizer, and
    differ only in transformer-block body initialization. The EAGLE-style
    DFlash variant preserves the pretrained drafter attention path at
    initialization using zero-initialized target-side projections. Bolded value indicates best performance. "math (ID)" means math in-domain.}
  \label{tab:dflash-controlled}
\end{table}

Under the original DFlash-style interface, warm-starting from the IDLM-style
diffusion checkpoint helps both in domain and out of domain, but the gain is
small: the warm start improves over the random-body baseline by at most
$0.07$ accepted tokens in any domain. The random-body baseline reaches
$5.96$ accepted tokens on math, but its four-domain OOD mean is only $1.64$;
the IDLM warm start improves this mean slightly to $1.68$.

We hypothesize that the small warm-start gain is partly caused by key/value
distribution drift in the original DFlash-style interface. Because the
drafter attention consumes target-derived key/value streams, the pretrained
IDLM body does not see the same key/value distribution it saw during diffusion
pretraining. This mismatch can reduce the amount of pretrained attention
computation that is reusable after fine-tuning, allowing the random-body
baseline to catch up quickly.

The EAGLE-style DFlash variant reduces this mismatch by preserving the
pretrained drafter attention path at initialization and adding target
conditioning through zero-initialized projections. It reaches an OOD mean
accepted length of $2.00$, substantially above the original DFlash-style warm-start
column. This suggests that, for diffusion drafters as well, warm-start
benefits are more likely to survive fine-tuning when the target interface
does not overwrite the pretrained drafter's attention pathway.

\enlargethispage{2\baselineskip}
\section*{Use of AI Assistants}

The authors used AI assistants for limited writing and coding support, including drafting language, editing for clarity, generating LaTeX snippets, and checking code or table formatting. 
All technical ideas, experimental design, implementation decisions, analyses, and final paper content were reviewed and verified by the authors, who remain fully responsible for the submission.
\end{document}